\documentclass[10pt,twocolumn,letterpaper]{article}

\usepackage{iccv}
\usepackage{times}
\usepackage{epsfig}
\usepackage{graphicx}
\usepackage{amsmath}
\usepackage{amssymb}

\usepackage[breaklinks=true,bookmarks=false]{hyperref}

\iccvfinalcopy 

\def\iccvPaperID{****} 

\ificcvfinal\fi

\begin{document}
\title{Multi-Stage HRNet: Multiple Stage High-Resolution Network for Human Pose Estimation}

\author{Junjie Huang, Zheng Zhu, Guan Huang\\
Institute of Automation, Chinese Academy of Sciences, Beijing.\\
{\tt\small zhengzhu@ieee.org}
}

\maketitle

\begin{abstract}

Human pose estimation are of importance for visual understanding tasks such as action recognition  and human-computer interaction. In this work, we present a Multiple Stage High-Resolution Network (Multi-Stage HRNet) to tackling the problem of multi-person pose estimation in images. Specifically, we follow the top-down pipelines and
high-resolution representations are maintained during single-person pose estimation. In addition, multiple stage network and cross stage feature aggregation are adopted to further refine the keypoint position. The resulting approach achieves promising results in COCO datasets. Our single-model-single-scale test configuration obtains 77.1 AP score in test-dev using publicly available training data. \footnote{technical report. Junjie Huang and Zheng Zhu contribute equally to this work.}
\end{abstract}

\section{Introduction}

Human pose estimation has witnessed a significant advance thanks to the development of deep learning. Motivated by practical applications in video surveillance \cite{li2019state}, human-computer interaction \cite{zhang2012microsoft, zhu2019human}, scene understanding \cite{li2018attention} and action recognition \cite{zhu2018action, zhu2019convolutional,zhu2018endto,zhu2018two}, researchers now switch focus from single person \cite{DeepPose,CPM,Hourglass,fastpose} to multi-person pose estimation in unconstrained environments \cite{DeepCut,DeeperCut,OpenPose,GooglePose,CPN,MSRAPose, MSPN, HRNet}. Even though research community has witnessed a significant advance, there are still challenging pose estimation problems in complex environments, such as occlusion, intense light and rare poses.

Multi-person pose estimation can be categorized into bottom-up \cite{DeepCut,DeeperCut,OpenPose} and top-down approaches \cite{GooglePose,CPN,MaskRCNN,MSRAPose}, where the latter becomes dominant participants in COCO benchmarks \cite{COCO}. Bottom-up architecture based methods first detect body parts and then associate corresponding body parts with specific human instances. Top-down approaches firstly detect and crop persons
from the image, then perform the single person pose estimation in the cropped person patches.

Recently, Multiple Stage Pose estimation Network (MSPN) \cite{MSPN} and High-Resolution Network (HRNet) \cite{HRNet} set new state-of-the-art performances in COCO keypoint benchmark. In this work, we design a Multiple Stage High-Resolution Network (Multi-Stage HRNet) by elegantly combining these two awesome approaches. Specifically, we follow the top-down pipelines and
high-resolution representations are maintained during single-person pose estimation. To further refine the keypoint position, multiple stage network and cross stage feature aggregation are adopted.

The proposed approach achieves promising performance in COCO keypoint benchmark. Without bells and whistles, Multi-Stage HRNet achieves 79.4 and 77.1 AP score on mini-validation and test-dev, with publicly available training data and single-model-single-scale test configuration. With simple and common test augmentation and model ensemble, we obtain 80.3 and 77.7 AP score in mini-validation and test-dev, respectively.

\section{Related Works}

\subsection{Single person pose estimation}
 Recently, single person pose estimation has been advanced rapidly for the development of deep convolution neural networks (CNN). DeepPose \cite{DeepPose} firstly tries to utilize CNN in pose estimation by directly regressing the $x$, $y$ coordinates of body parts. More recently, researchers choose to regress heatmaps, where each peak stands for a body part. With the continuous work of research community, novel architectures such as CPM \cite{CPM} and Stacked Hourglass \cite{Hourglass} are proposed to achieve better results.

\subsection{Multi-person pose estimation}
Different from single pre-located person, multi-person pose estimation can be categorized into bottom-up \cite{DeepCut,DeeperCut,OpenPose} and top-down approaches \cite{GooglePose,CPN,MaskRCNN,MSRAPose,MSPN,HRNet}.

\paragraph{bottom-up} Bottom-up architecture based methods adopt a different work flow, which first detect body parts and then associate corresponding body parts with specific human instances. The typical methods are DeepCut \cite{DeepCut} and DeeperCut \cite{DeeperCut}, the former adopts an integer linear programming(ILP) based method and the later improves DeepCut via utilizing image-conditioned pairwise terms. Cao et al. \cite{OpenPose} predict heatmaps of body parts and a set of 2D vector fields of part affinities and parse them by greedy inference to generate the final results.

\paragraph{Top-down}
CPN \cite{CPN} is the leading method on COCO 2017
keypoint challenge. It involves skip layer feature concatenation and an online hard keypoint mining step. \cite{MSRAPose} adopts FPN-DCN as the human detector and adds a few deconvolutional layers on single-person pose estimation network to improve the performance. Besides, Mask R-CNN \cite{MaskRCNN} builds an end-to-end framework and yields an impressive performance. Recently, HRNet \cite{HRNet} and MSPN \cite{MSPN} set new state-of-the-art results on COCO keypoint detection task. HRNet maintains high-resolution
representations through the whole single person pose estimation by connecting the multi-resolution subnetworks in parallel. MSPN propses a multi-stage pose estimation
network with feature aggregation across different stages and coarse-to-fine supervision strategy.


\section{Multi-Stage HRNet}

\subsection {Overall framework}
\begin{figure}[tp]
\centering
\includegraphics[width=0.5\textwidth]{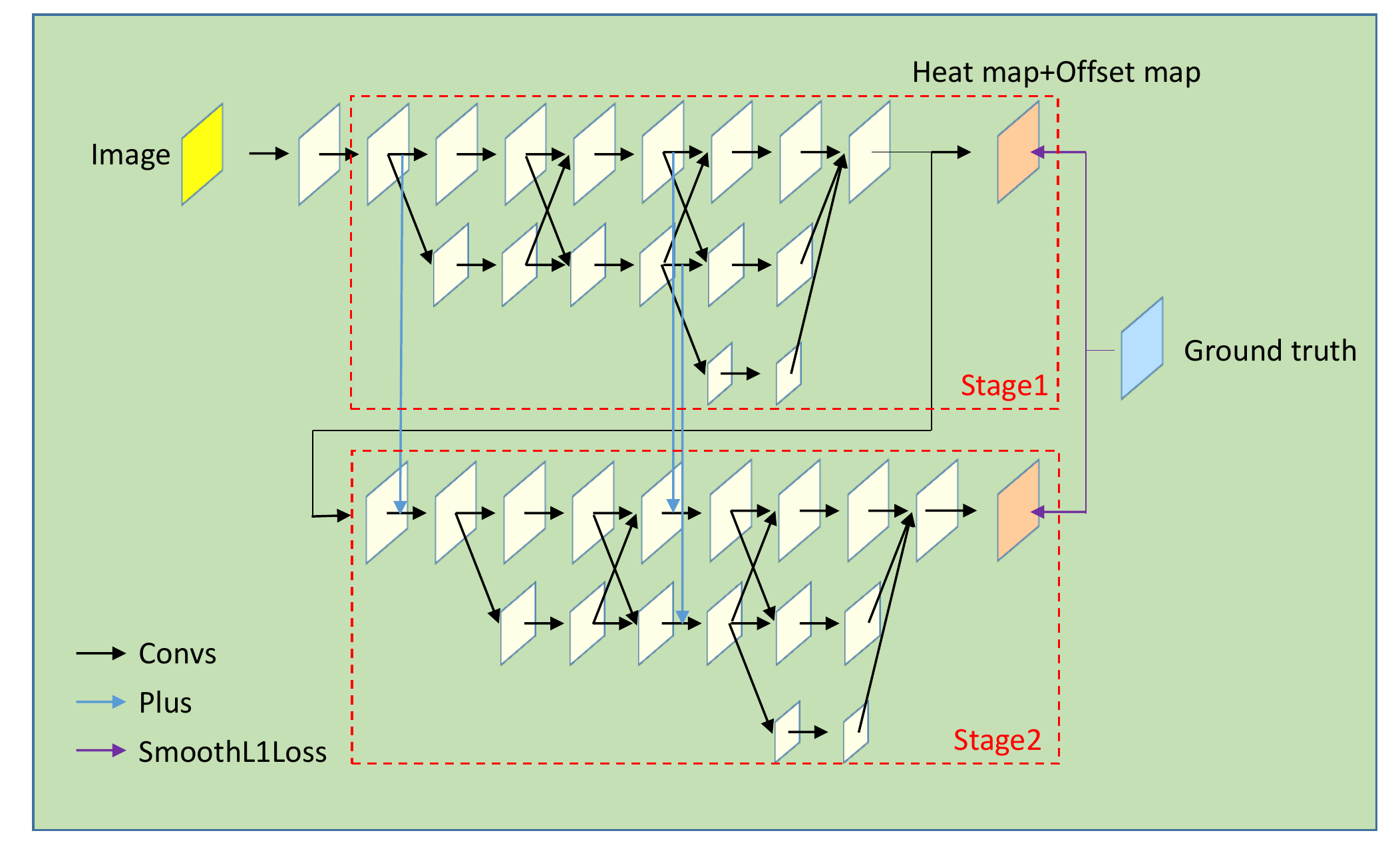}
\caption{Overall framework of Multi-Stage HRNet.
}
\label{framework}
\end{figure}

The overall framework of Multi-Stage HRNet is illustrated in Figure \ref{framework}. In each stage, HRNet is adopted as backbone, which is potentially more accurate and spatially
more precise than high-to-low resolution networks \cite{MSRAPose,CPN,Hourglass}. Following \cite{HRNet}, our approach maintains high-resolution
representations through the whole single person pose estimation by connecting the multi-resolution subnetworks in parallel. Besides, repeated multi-scale fusions are performed, which makes each of the high-to-low resolution representations receive information from other parallel representations. Since estimating the keypoint positions by single network is difficult, multiple high-resolution networks are cascaded to refine the pose results. Feature maps yielded by each high resolution block are delivered to the corresponding positions of the next stage by performing plus operation. Different from 4 stages used in MSPN, here we adopt 2 stages network due to training efficiency and GPU memory. Following \cite{GooglePose}, we finally decode the heat map and offset map of the last stage to position the keypoints.

\subsection {Training details}

The proposed Multi-Stage HRNet is implemented using MXNet framework. We start this work from August of 2019.
There are totally 4 machines to train models, and each machine is equipped with 8 Titan RTX GPUs (24G). The multiple stage weight is initialized using single stage weight from classification task.
 The Adam solver with fp32 is used for training (We use fp16 to accelerate training speed at first, but it is unstable under some epochs). The initial learning rate is set to 0.001, and reduced by a factor of 10 at 110 and 140 epochs, respectively. Most model is trained with 150 epochs, and epochs of some model varies due to time limitations. Training data consists of 4 publicly available datasets, including COCO \cite{COCO}, MPII \cite{MPII}, AI Challenger \cite{AIC}, CrowdPose \cite{CrowdPose}. The image/instance numbers of these datasets are listed in Table \ref{training_data}.
For data argumentation, random rotation ($[-30^{\circ}, 30^{\circ}]$), horizontal flipping,
random resizing ($[0.8, 1.2]$), half body cropping are utilized.

\begin{table}[!ht]
\footnotesize
  \centering
  \caption{Details about training dataset, numbers are calculated by images containing pose annotations.}
\begin{tabular}{c|c|c}
\hline
\bf dataset & image number & instance number\\\hline
\bf COCO & 56599 & 149760 \\\hline
\bf MPII &  17408& 28800 \\\hline
\bf AI Challenger & 209888 & 377856 \\\hline
\bf CrowdPose & 12000 & 42624 \\\hline
\end{tabular}
   \label{training_data}
\end{table}

\subsection {Test details}

For person detector, we use the HTC \cite{chen2019hybrid} with multi-scale test. The 80-class and person AP on mini-validation are 52.9 and 65.1, respectively. In ablation study, we adopt single-model-single-scale configuration with flipping strategy. No test argumentation or ensemble are performed.
For finally ensemble, three 2-stage (i.e Multi-Stage HRNet-W48*2) are utilized and results are obtained by averaging the position in images.
For test argumentation in ensemble, rotation ($\pm20^{\circ}$) and multi-scale (3 scales) are used.

\section{Experiments}

In this section, we report the preliminary results on COCO mini-validation and test-dev dataset. It is noting that results may update according to further experiments.

\subsection {Ablation study}
In this section, the ablation study of Multi-Stage HRNet is performed and results are listed in Table \ref{Ablation}.

\paragraph{Input size} In mini-validation set and HRNet-w32 backbone, the AP increase from 76.3 to 77.6 when input size is from $256\times192$ to $384\times288$. For HRNet-w48 backbone, the AP steadily increases when input size gets larger. The performance in test-dev set is similar with mini-validation.

\paragraph{Backbones} Larger backbones always bring better performance. In mini-validation set and $256\times192$ input size, the AP of HRNet-w32 and HRNet-w48 backbone are 76.3 and 76.9, respectively. In test-dev, the performance is 73.9 and 74.3, respectively.

\paragraph{Multiple stages} We validate the effectiveness of multiple stages with HRNet-w48 backbone and $512\times384$ input size. In mini-validation and test-dev set, the improvements are 0.4 and 0.3, respectively.

\paragraph{Training data and ensemble} More training data could boost the representation and generalization of deep learning models. In this work, we only utilize the publicly available pose data to train the Multi-Stage HRNet, which is illustrated in Table \ref{training_data}. As shown in Table \ref{Ablation}, additional data could boost mini-validation and test-dev set by 1.2 and 1.3 AP, respectively. Finally, ensemble brings 0.9 and 0.6 AP for mini-validation and test-dev set.

\begin{table*}[!ht]
\footnotesize
  \centering
  \caption{Ablation study of Multi-Stage HRNet in mini-validation and test-dev set.}
\begin{tabular}{c|c|c|c|c|c|ccccc}
\hline
 Stages & Backbone &Input size & training data &ensemble & dataset & AP & $AP^{50}$ & $AP^{75}$ & $AP^{M}$ & $AP^{L}$\\\hline
 1 & HRNet-w32 & $256 \times 192$ & COCO & No & mini-val & 76.3 & 92.6 & 83.7 & 73.8 & 81.9\\\hline
 1 & HRNet-w32 & $384 \times 288$ & COCO & No & mini-val & 77.6 & 91.4 & 82.9 & 73.9 & 83.7\\\hline
 1 & HRNet-w48 & $256 \times 192$ & COCO & No & mini-val & 76.9 & 90.8 & 82.3 & 73.5 & 83.2\\\hline
 1 & HRNet-w48 & $512 \times 384$ & COCO & No & mini-val & 77.8 & 91.6 & 83.3 & 74.3 & 84.3\\\hline
 2 & HRNet-w48 & $512 \times 384$ & COCO & No & mini-val & 78.2 & 91.8 & 83.7 & 74.7 & 85.1\\\hline
 2 & HRNet-w48 & $512 \times 384$ & all  & No & mini-val & 79.4 & 92.6 & 85.3 & 75.3 & 86.0 \\\hline
 2 & HRNet-w48 & $512 \times 384$ & all  & Yes & mini-val & 80.3 & 93.1 & 86.4 & 75.9 & 87.2 \\\hline

 \hline
 1 & HRNet-w32 & $256 \times 192$ & COCO & No & test-dev & 73.9 & 91.8 & 80.8 & 70.3 & 79.9 \\\hline
 1 & HRNet-w32 & $384 \times 288$ & COCO & No & test-dev & 75.0 & 92.1 & 83.1 & 71.6 & 81.3\\\hline
 1 & HRNet-w48 & $256 \times 192$ & COCO & No & test-dev & 74.3 & 92.0 & 81.4 & 70.8 & 80.4\\\hline
 1 & HRNet-w48 & $512 \times 384$ & COCO & No & test-dev & 75.5 & 92.2 & 83.8 & 71.8 & 81.6\\\hline
 2 & HRNet-w48 & $512 \times 384$ & COCO & No & test-dev & 75.8 & 92.3 & 84.0 & 72.1 & 81.6\\\hline
 2 & HRNet-w48 & $512 \times 384$ & all  & No & test-dev & 77.1 & 92.5 & 84.2 & 73.8 & 82.9\\\hline
  2 & HRNet-w48 & $512 \times 384$ & all  & Yes & test-dev & 77.7 & 93.0 & 84.8 & 74.1 & 83.7\\\hline
\end{tabular}
   \label{Ablation}
\end{table*}

\subsection {Comparison with other methods}

In this section, the performance of proposed method is compared with single-model methods on COCO test-dev set. As illustrated in Table \ref{table:coco_test_dev}, our 2-atage HRNet obtains 75.8 and 77.1 with COCO and all publicly available data respectively, which is very promising. It is worth noting that HRNet use smaller backbone and less training data.

	\begin{table*}[t]
		\caption{Comparisons on the COCO test-dev set with most single-model configuration.}
	\centering
			\label{table:coco_test_dev}
			\footnotesize
            \begin{tabular}{l|l|c|c|c|lllll}
				\hline
				Method &Backbone& Input size & \#Params & GFLOPs&
				$\operatorname{AP}$ & $\operatorname{AP}^{50}$ & $\operatorname{AP}^{75}$ & $\operatorname{AP}^{M}$ & $\operatorname{AP}^{L}$ \\
				\hline
				Mask-RCNN~\cite{MaskRCNN} & ResNet-50-FPN& $-$ &$-$& $-$
				& $63.1$ & $87.3$&$68.7$&$57.8$&$71.4$\\
				G-RMI~\cite{GooglePose} & ResNet-101 & $353\times257$ &$42.6$M& $57.0$
				&$64.9$ & $85.5$&$71.3$&$62.3$&$70.0$\\
				G-RMI + extra data~\cite{GooglePose} & ResNet-101 & $353\times257$ &$42.6$M& $57.0$
				&$68.5$ & $87.1$&$75.5$&$65.8$&$73.3$\\
				CPN~\cite{CPN} & ResNet-Inception& $384\times288$ &$-$& $-$
				& $72.1$ & $91.4$&$80.0$&$68.7$&$77.2$\\
				RMPE~\cite{fang2017rmpe} & PyraNet & $320\times256$ &$28.1$M& $26.7$
				&$72.3$ & $89.2$&$79.1$&$68.0$&$78.6$\\
				CPN (ensemble)~\cite{CPN} & ResNet-Inception& $384\times288$ &$-$& $-$
				&$73.0$ & $91.7$&$80.9$&$69.5$&$78.1$\\
				SimpleBaseline~\cite{MSRAPose} & ResNet-152&$384\times288$  &$68.6$M& $35.6$
				&${73.7}$ & ${91.9}$&${81.1}$&${70.3}$&${80.0}$\\
				HRNet-W$32$~\cite{HRNet}& HRNet-W$32$& $384\times 288$ &$28.5$M&$16.0$ &$74.9$&$92.5$&$82.8$&$71.3$&$80.9$\\
				HRNet-W$48$~\cite{HRNet} & HRNet-W$48$& $384\times 288$ &$63.6$M& $32.9$
				& $75.5$&$92.5$&$83.3$&$71.9$&$81.5$\\
				HRNet-W$48$ + extra data ~\cite{HRNet}& HRNet-W$48$& $384\times 288$ &$63.6$M& $32.9$
				& $77.0$&$92.7$&$84.5$&$73.4$&$83.1$\\
				\hline
				Ours & HRNet-W$48*2$& $512\times 384$              & $118.2$M &61.5 & 75.8 & 92.3 & 84.0 & 72.1 & 81.6\\
				Ours + extra data & HRNet-W$48*2$& $512\times 384$ & $118.2$M &61.5 & 77.1 & 92.5 & 84.2 & 73.8 & 82.9\\
				\hline
			\end{tabular}
	\end{table*}

We also compare ensemble results with methods in COCO leaderboard. As shown in Table \ref{Comparisons_leaderboard}, our AP is 77.7 which could rank second. The ranked first Megvii (Face++) use private data for training.

\begin{table}[!ht]
\footnotesize
  \centering
  \caption{Comparison with methods in COCO leaderboard.}
\begin{tabular}{c|c|c|c}
\hline
\bf Methods  & Backbone & training data & AP \\\hline
\bf Ours                & HRNet &Public&77.7 \\\hline
\bf Megvii (Face++)     & ResNet &Public+Private & 78.1\\\hline
\bf MSRA                & ResNet &Public & 76.5\\\hline
\bf The Sea Monsters    & ResNet &Public+Private & 75.9\\\hline
\bf KPLab               & ResNet       &- & 75.1\\\hline
\bf DGDBQ               & ResNet       &Public & 74.9\\\hline
\bf ByteDance-SEU       & ResNet       &Public & 74.2\\\hline

\end{tabular}
   \label{Comparisons_leaderboard}
\end{table}

\section{Conclusion and future work}

In this work, we present Multi-Stage HRNet to tackling the problem of multi-person pose estimation in images. Following the top-down pipelines, high-resolution representations are maintained during single-person pose estimation, and multiple stage network are adopted to further refine the keypoint position. Without bells and whistles, Multi-Stage HRNet achieves 79.4 and 77.1 AP score on mini-validation and test-dev, with and publicly available training data and single-model-single-scale test configuration.
Due to time and GPU resources limitation, we currently adopt simple and common data augmentation during training and test. There also may be some misalignments for flip, rotation and resize.
Future work may fix these misalignments and explore special augmentation strategies. Another future work may combine Multi-Stage HRNet with tracking strategy \cite{UCT, SiamRPN, DaSiamRPN,FlowTrack,li2019state} for pose tracking tasks \cite{Posetrack,zhang2019fastpose,zhang2019exploiting}.

{\footnotesize
 \bibliographystyle{ieee} \bibliography{egbib}}

\end{document}